# PersonaDrift: *A Benchmark for Temporal Anomaly Detection in Language-Based Dementia Monitoring*

Joy Lai
Institute of Biomedical Engineering
University of Toronto
Toronto, Canada
joy.lai@mail.utoronto.ca

Alex Mihailidis
Institute of Biomedical Engineering
University of Toronto
Toronto, Canada
alex.mihailidis@utoronto.ca

*Abstract* — People living with dementia (PLwD) often show gradual shifts in how they communicate, becoming less expressive, more repetitive, or drifting off-topic in subtle ways. While caregivers may notice these changes informally, most computational tools are not designed to track such behavioral drift over time. This paper introduces PersonaDrift, a synthetic benchmark designed to evaluate machine learning and statistical methods for detecting progressive changes in daily communication, focusing on user responses to a digital reminder system.

PersonaDrift simulates 60-day interaction logs for synthetic users modeled after real PLwD, based on interviews with caregivers. These caregiver-informed personas vary in tone, modality, and communication habits, enabling realistic diversity in behavior. The benchmark focuses on two forms of longitudinal change that caregivers highlighted as particularly salient: flattened sentiment (reduced emotional tone and verbosity) and off-topic replies (semantic drift). These changes are injected progressively at different rates to emulate naturalistic cognitive trajectories, and the framework is designed to be extensible to additional behaviors in future use cases.

To explore this novel application space, we evaluate several anomaly detection approaches, unsupervised statistical methods (CUSUM, EWMA, One-Class SVM), sequence models using contextual embeddings (GRU + BERT), and supervised classifiers in both generalized and personalized settings. Preliminary results show that flattened sentiment can often be detected with simple statistical models in users with low baseline variability, while detecting semantic drift requires temporal modeling and personalized baselines. Across both tasks, personalized classifiers consistently outperform generalized ones, highlighting the importance of individual behavioral context.

By providing a structured, longitudinal, and caregiver-informed benchmark, PersonaDrift lays critical groundwork for next-generation NLP systems that can track subtle, personalized communication changes over time. It supports the development of ethical, privacy-aware monitoring tools that are responsive to the evolving needs of PLwD, helping caregivers and clinicians detect early signs of change and provide timely support.

## I. INTRODUCTION

Language reflects cognitive state in subtle but meaningful ways, and in people living with dementia (PLwD), patterns of communication often shift over time [1], [2]. Responses may become flatter, more repetitive, or increasingly off-topic [3], [4]. While caregivers may notice these changes informally, current computational systems rarely support structured, long-term tracking of such drift [5]. As part of ongoing work on conversational reminder systems for PLwD, we interviewed PLwD and their caregivers [6], [7]. A recurring request was the ability to monitor changes in how users respond over time, supporting early detection of decline or adaptation of care.

Despite the promise of natural language processing (NLP) for behavioral monitoring in dementia care, most existing tools are not designed for progressive, personalized drift [5], [8]. Clinical datasets such as DementiaBank provide valuable language samples, but their structured, short-form interactions do not reflect the longitudinal, spontaneous conversations that occur in home settings [5], [9], [10]. There is a clear gap: systems need to detect emerging trends in communication style, not for diagnostic labeling, but to support ongoing monitoring and caregiver insight.

We introduce PersonaDrift, a synthetic benchmark designed specifically for this task. It simulates 60-day interaction logs between reminder systems and synthetic users modeled on real-world PLwD, based on interviews capturing their routines, tones, modalities, and response styles [6], [11]. The benchmark focuses on two behavioral shifts that caregivers identified as relevant and tractable for early monitoring: flattened sentiment (reduced expressiveness) and off-topic replies (semantic drift). These anomalies are injected progressively at varying speeds to reflect realistic cognitive change.

To ensure the benchmark is both challenging and representative, we evaluate a diverse set of detection methods, from simple statistical baselines to context-aware sequence models, and supervised classifiers in both personalized and generalized forms. These experiments are not intended to propose a new model, but rather to validate the benchmark's utility and expose challenges in this application space, such as threshold instability and the critical role of user-specific baselines.

This work was supported by AGE-WELL NCE (RIS Fund Number 499052) and the University of Toronto.

PersonaDrift provides a controlled, extensible framework for studying linguistic drift in cognitively relevant settings. It enables reproducible evaluation of NLP methods that aim to support PLwD through unobtrusive, adaptive monitoring, laying the groundwork for privacy-preserving, caregiver-aligned systems that respond to real communication needs.

## II. RELATED WORK

### A. Language Change in Dementia

Changes in language use are well-documented in dementia and related neurodegenerative conditions. Prior studies have identified early signs such as reduced lexical diversity, increased pronoun use, topic repetition, and semantic incoherence as indicators of cognitive decline [3], [4]. Existing datasets like DementiaBank and CUBOld have been central to this work, offering annotated speech transcripts from structured interviews with PLwD [9], [10]. However, these datasets are typically episodic and lack the continuity or ecological grounding needed to study how communication patterns evolve in real-world, daily-life contexts. As a result, they offer limited utility for evaluating systems designed to monitor subtle, progressive behavioral drift over time.

### B. Behavioral Monitoring in Digital Health

Natural language is increasingly recognized as a potential signal in digital health applications, including tools for cognitive monitoring and remote caregiving [8], [12]. Prior work has focused on passive sensing of behavior through mobility patterns, device usage, or acoustic features, but few systems leverage longitudinal language data [13], [14]. In our ongoing work, we are developing a home-based reminder system that allows caregivers to schedule prompts and monitor responses from PLwD remotely [6], [7]. The system supports different response modalities (e.g., typed or spoken) and includes lightweight follow-up questions to check task completion or general well-being. Initial interviews and usability testing with PLwD and caregivers motivated the need for behavioral monitoring features that could detect gradual communication changes and help inform support needs. PersonaDrift was designed to benchmark potential detection methods for this application.

### C. Dialogue Simulation and Persona Modeling

Large language models (LLMs) are increasingly used to simulate user behavior in domains such as healthcare, education, and human-AI interaction [15], [16], [17], [18]. Persona-based modeling allows for variation in tone, style, and response habits, enabling targeted evaluation of dialogue systems [19], [20]. However, most prior work assumes static user traits and does not simulate progressive behavioral change [21]. Some simulation frameworks include stochastic variability or task-specific intent drift, but few are designed to capture cognitively relevant, gradual degradation in open-ended conversation [22], [23]. This limits their applicability in domains like elder care, where adaptation over time is critical.

### D. Conversational Anomaly Detection

Anomaly detection in NLP has typically focused on short-term failures, such as hallucinated outputs, abrupt topic shifts, or turn-level incoherence [24], [25]. These are often detected using model confidence scores, classifier thresholds, or rule-based filters. While effective for identifying discrete issues, such approaches are poorly suited for detecting gradual or user-specific anomalies, such as those caused by cognitive drift. Capturing these subtler changes requires temporally grounded models that can compare responses against individualized linguistic baselines.

### E. Temporal and Personalized Modeling

Temporal modeling and personalization have shown success in emotion recognition, user state tracking, and adaptive dialogue systems [26], [27]. Sequence models like RNNs and attention-based architectures can learn long-range dependencies, while personalized models often outperform general ones in high-variability domains [28], [29], [30]. Yet, few studies combine both in the context of long-term, health-related dialogue monitoring. Reminder-based systems, which involve repeated prompts in daily routines, offer a structured setting for this type of modeling but lack existing benchmarks that simulate progressive linguistic change grounded in individual user traits.

## III. BENCHMARKING PIPELINE

PersonaDrift simulates longitudinal interactions with a digital reminder system for PLwD. Each synthetic user (persona) reflects real routines, communication styles, methods (typing or voice-to-text), and tone patterns drawn from interviews with PLwD [7]. The pipeline consists of four stages: routine schedule generation, random event injection, response generation, and anomaly injection. It produces realistic, time-stamped logs for evaluating behavioral drift detection methods.

### A. Routine Simulation

Personas follow weekly schedules of recurring reminders (e.g., medication, hygiene, check-ins), based on real user routines. For each reminder, the system determines acknowledgment using a delay model:

$$t_{ack} = t_{reminder} + \delta_r \qquad (1)$$

Here, $\delta_r$ is drawn from a persona-specific delay distribution, either normal $N(\mu_r, \sigma_r)$ or triangular $Tri(a_r, b_r, c_r)$, reflecting individualized response timing. A response is logged only if $t_{ack}$ falls within a valid window and passes a Bernoulli trial based on location-specific response likelihood. This yields a structured log of acknowledged and missed reminders over a 60-day simulation period.

PersonaDrift was designed as an evaluation benchmark rather than a large-scale training corpus. For anomaly detection, scale requirements differ by method: rule-based baselines and embedding-based similarity methods can operate effectively with smaller curated sets, while neural architectures benefit from larger corpora. Our dataset size is therefore sufficient to evaluate the phenomena of interest, while leaving room for future expansion. This balance of feasibility and ecological grounding reflects deliberate design rather than scarcity.

## B. Random Event Injection

To introduce variability, random reminders are sampled daily using a persona-specific frequency and type distribution. Event types (e.g., appointments, check-ins, household tasks) are selected via custom weighted multinomial sampling and assigned plausible times and locations. GPT-based models generate reminder text using few-shot prompting from the persona's prior messages, preserving tone, modality (typed/spoken), and style. These events follow the same response modeling as routine ones.

## C. Response Generation

For every acknowledged reminder, the system generates a user response conditioned on persona attributes. Large language models produce utterances aligned with the user's expressiveness and modality, e.g., a terse voice user might say, "Yeah, I did," while a verbose typed user might write a detailed explanation. This stage transforms event logs into full conversational episodes, enabling drift-aware analysis.

## D. Anomaly Choice

Our anomaly selection was grounded in qualitative caregiver interviews, where conversational drift and emotional flattening emerged as the most salient and frequently mentioned communication breakdowns. While other anomalies are important in clinical and linguistic research, they did not emerge as primary caregiver concerns in our interviews and were therefore not prioritized in this initial release.

## E. Anomaly Injection

To simulate progressive decline, the benchmark injects two types of anomalies: flattened sentiment (reduced emotional tone and verbosity) and off-topic replies (semantic drift from the prompt). Anomalies are introduced over a randomly selected window and progress at one of three speeds: slow (15–20 days), medium (10–14), or fast (6–8). This approach is inspired by drift modeling techniques used in synthetic anomaly detection benchmarks, where anomalies are injected at varying intensities and progression speeds to simulate real-world behavior [31], [32].

Severity scales over time using:

$$Severity(t) = \min\left(3, \left\lfloor \alpha \cdot \frac{t-t_{start}}{d} \right\rfloor + 1\right) \quad (2)$$

where $t$ is the current day, $t_{start}$ is the anomaly onset, $d$ is the duration, and $\alpha$ controls speed (e.g., $\alpha = 2$ for slow, 5 for fast). Not every response within the window is affected; severity influences the likelihood and nature of modification. Responses are altered using structured prompts tailored to each persona's style. Mild anomalies may retain partial topical relevance, while severe ones introduce unrelated or affectively flat replies. All changes are logged with metadata, including type, severity, and affected days, to support downstream filtering and evaluation.

## IV. EVALUATION OF BENCHMARK PIPELINE

We evaluate the benchmark using a range of baseline methods across both anomaly types, flattened sentiment and off-topic replies. The goal is not to optimize performance, but to assess benchmark difficulty and surface key challenges for longitudinal, personalized monitoring.

### A. Flattened Sentiment Detection

Flattened sentiment refers to a gradual reduction in emotional tone, subjectivity, and verbosity in user responses. This detection task is framed as a binary classification problem, where each response is labeled as either typical or anomalous based on simulated affective flattening.

We evaluate three unsupervised methods:

- Cumulative Sum Control Chart (CUSUM) [33]
- Exponentially Weighted Moving Average (EWMA) [34]
- One-Class Support Vector Machine (One-Class SVM) [35]

Each method operates on a shared feature set designed to capture surface-level affective tone while minimizing dependence on prompt content. The features include:

- Valence Aware Dictionary for Sentiment Reasoning (VADER) sentiment score
- TextBlob polarity
- TextBlob subjectivity
- Normalized response length

#### 1) CUSUM

CUSUM tracks sustained negative shifts from a user's baseline tone. At each step, a cumulative deviation score $S_n$ is updated as shown in (3):

$$S_n = \min(0, S_{n-1} + x_n - \mu + k) \quad (3)$$

where $x_n$ is the current tone feature, $\mu$ is the baseline mean (first 10 responses), and $k$ controls sensitivity. Anomaly flags are raised when $S_n$ drops below a fixed threshold, enabling detection of gradual affective flattening.

#### 2) EWMA

EWMA smooths tone features to emphasize recent values, updating the signal as:

$$S_n = \lambda x_n + (1 - \lambda) S_{n-1} \quad (4)$$

where $x_n$ is the current tone feature and $\lambda \in (0,1)$ controls responsiveness. Anomalies are flagged when standardized residuals (z-scores) exceed a fixed threshold. EWMA reacts quickly to short-term deviations but is sensitive to natural variability, particularly in expressive users.

#### 3) One-Class SVM

One-Class SVM models the distribution of normal responses in tone-feature space using a radial basis function (RBF) kernel. Outliers are classified as anomalies. This approach can capture non-linear tone shifts but is sensitive to baseline noise and limited training data.

Performance is evaluated using two key metrics:

- **F1 Score**: Balances precision and recall:

$$F1 = \frac{2 \cdot Precision \cdot Recall}{Precision + Recall} \quad (5)$$

- **Average Detection Delay**: Measures the mean time (in days) between anomaly onset and first correct detection, calculated only for cases with successful identification.

Evaluations are performed per persona to assess method robustness across different communication styles, modalities, and baseline tone stability.

*B. Off-Topic Detection*

Off-topic responses, marked by semantic drift away from the intended prompt, are more variable and context-dependent than affective flattening, making them harder to detect. These shifts often occur gradually and may not exhibit clear surface features, requiring context-aware methods. We evaluate detection performance using F1 score and area under the receiver operating characteristic curve (ROC AUC), capturing both classification accuracy and anomaly ranking.

For the CUSUM and One-Class Support Vector Machine (SVM) baselines, we use a compact set of semantic features: (1) semantic drift (cosine distance from expected response), (2) similarity to prompt, (3) similarity to the reminder message, and (4) keyword overlap. In contrast, the Gated Recurrent Unit (GRU) and personalized versus generalized classifiers operate over contextualized BERT (Bidirectional Encoder Representations from Transformers) embeddings of user responses.

Ground-truth labels for off-topic responses are derived from controlled anomaly injections within the simulated logs.

*1) CUSUM*

We apply a modified CUSUM algorithm to the semantic drift signal (cosine distance from expected response). Rather than using a fixed threshold, we select a user-specific value from a grid search over 50 candidates ($\theta \in [0.01, 0.5]$). Anomaly detection is defined as:

$$S_t = max(0, S_{t-1} + x_t - \theta) \quad (6)$$

An event is flagged when $S_t > \theta$. ROC AUC is calculated on raw drift scores to evaluate ranking. While interpretable, performance is constrained by baseline sensitivity.

*2) GRU-Based Sequence Modeling*

To model long-range coherence, we train a gated recurrent unit (GRU) model on **BERT embeddings** of each user's responses. Given a sliding window of size $W = 5$, the GRU predicts the next response embedding:

$$\hat{x}_t = GRU(x_{t-W+1}, \ldots, x_t) \quad (7)$$

Prediction error is measured by cosine distance:

$$error_t = 1 - cos(\hat{x}_t, x_t) \quad (8)$$

Anomalies are flagged when error exceeds the 95th percentile of errors from presumed-normal data. This approach captures personalized semantic trajectories and can identify subtle, context-specific drift.

*3) One-Class SVM on Semantic Features*

We apply a One-Class Support Vector Machine (SVM) to the previously described semantic features. Inputs are z-normalized, and the model is trained on each user's first 10 days of responses, assuming typical behavior. An anomaly is flagged if the test point lies outside the learned boundary:

$$f(x) = \sum_i \alpha_i K(x, x_i) - \rho \quad (9)$$

where $K(\cdot,\cdot)$ is the RBF kernel.

*C. Supervised Detection: Personalized vs. General*

This comparison enables direct evaluation of personalization's value in detecting off-topic conversational drift, especially in cognitively relevant, temporally structured dialogue settings. Both approaches use the same input features, BERT embeddings of user responses, but differ in inductive bias: personalized models capture user-specific patterns over time, while generalized models aim to detect shared semantic anomalies across a population. To assess the impact of personalization, we compare two supervised learning strategies:

*1) Personalized Setting:*

A separate logistic regression model is trained per user on their own labeled responses, enabling adaptation to individual semantic norms. The model computes

$$P(y = 1|x) = \sigma(\mathbf{w}^T \mathbf{x} + b) \quad (10)$$

where $x \in \mathbb{R}^d$ is the BERT embedding, $\mathbf{w}$ are learned weights, $b$ is a bias term, and $\sigma$ is the sigmoid activation. Models are trained with **balanced class weights** to address label imbalance and evaluated using stratified train-test splits on the same user.

*2) Generalized Setting:*

A leave-one-user-out approach is used, training on all users except one and testing on the held-out user. Training data is class-balanced and downsampled to simulate zero-shot deployment. This setup evaluates the model's ability to generalize across users without personalization.

## V. RESULTS

We evaluate model performance across eight synthetic personas, each reflecting real-world behavior patterns observed in PLwD. These personas vary in tone, verbosity, modality, and daily habits. Some exhibit time-dependent variability, for example, reduced coherence or affect later in the day, mirroring phenomena like sundowning. Full persona details are provided in Table I.

TABLE I. SUMMARY OF RESPONSE SETTINGS FOR EACH PERSONA

| Persona | Tone | Style | Expressiveness | *Mode* |
|---|---|---|---|---|
| 1 | Brief, polite | Short, neutral | Low | Typed |
| 2 | Concise, respectful | Short, routine-compliant | Low | Typed |
| 3 | AM: Kind, concise PM: Hesitant, nervous | AM: Short PM: Wordier responses | Moderate | Voice |
| 4 | Warm, responsive | Short, affirmative | Moderate | Typed |
| 5 | Calm, responsive | Short, friendly | Moderate | Voice |

| | | | | | | |
|---|---|---|---|---|---|---|
| 6 | Friendly, positive | Casual | High | Typed | | |
| 7 | Indifferent, cold | Short | Low | Voice | | |
| 8 | AM: Warm, polite PM: Irritable, flat | Short and simple | Moderate | Voice | | |

## A. Flattened Sentiment Detection

We evaluated flattened sentiment detection using three unsupervised methods, CUSUM, EWMA, and One-Class SVM, across eight personas and three progression speeds (fast, medium, slow). Performance is reported using F1 score and average detection delay (Tables II–IV).

**CUSUM** consistently delivered strong performance across all progression speeds. Most personas achieved F1 scores above 0.98 with negligible detection delay, particularly Personas 1–6. These users exhibit stable communication patterns: typed users like Personas 1 and 2 (brief, low expressiveness), and voice users like Personas 3 and 5 (moderate expressiveness) allowed CUSUM to effectively detect tone flattening over time. Even in slow progression, CUSUM maintained >0.96 F1 for nearly all personas.

**EWMA** was more sensitive to natural tone variability. It worked best for high or moderate expressiveness personas with relatively smooth trajectories, like Persona 6 (typed, friendly, high expressiveness) and Persona 5 (voice, friendly). For example, Persona 6 showed consistently high EWMA F1 scores (0.94–0.99) across all conditions. However, in users with expressive or time-varying tone, like Persona 8 (voice, polite in AM, flat in PM), EWMA often produced lower scores and delayed detections, especially under medium and slow progression (F1 = 0.32 in slow).

**SVM** performance fell between CUSUM and EWMA. It performed well for typed users with moderate expressiveness (e.g., Persona 4 and 6) and also matched CUSUM's F1 in some cases (e.g., Persona 6 at 0.987). However, it struggled with time-sensitive or low-affect voice users like Persona 7 (voice, indifferent), where performance was poor (F1 = 0.203 in fast, 0.338 in slow) and delay was high (2.46 days in fast condition).

Overall, **CUSUM** was the most robust method, excelling in users with consistent tone regardless of mode or expressiveness. **EWMA** and **SVM** were more variable and showed limitations in handling natural fluctuations, especially in voice-based or time-shifting personas like 3, 7, and 8. These results suggest that lightweight statistical detectors can be effective but must be tailored to user style and variability.

TABLE II. RESULTS OF FAST FLATTENED SENTIMENT DETECTION

| Persona | CUSUM | | EWMA | | SVM | |
|---|---|---|---|---|---|---|
| | F1 | Delay | F1 | Delay | F1 | Delay |
| 1 | 1.000 | 0.000 | 0.510 | 0.390 | 0.566 | 0.480 |
| 2 | 0.984 | 0.000 | 0.359 | 3.420 | 0.667 | 0.500 |
| 3 | 1.000 | 0.000 | 0.082 | 0.750 | 0.907 | 0.160 |
| 4 | 0.989 | 0.000 | 0.291 | 0.110 | 0.956 | 0.020 |
| 5 | 1.000 | 0.000 | 0.429 | 0.000 | 0.927 | 0.050 |
| 6 | 1.000 | 0.000 | 0.973 | 0.000 | 0.973 | 0.000 |
| 7 | 0.873 | 0.190 | 0.536 | 0.670 | 0.203 | 2.460 |
| 8 | 1.000 | 0.000 | 0.242 | 0.000 | 0.816 | 0.270 |

TABLE III. RESULTS OF MEDIUM FLATTENED SENTIMENT DETECTION

| Persona | CUSUM | | EWMA | | SVM | |
|---|---|---|---|---|---|---|
| | F1 | Delay | F1 | Delay | F1 | Delay |
| 1 | 0.982 | 0.020 | 0.328 | 0.720 | 0.444 | 0.710 |
| 2 | 0.958 | 0.000 | 0.571 | 0.090 | 0.750 | 0.330 |
| 3 | 1.000 | 0.000 | 0.118 | 0.000 | 0.843 | 0.050 |
| 4 | 0.991 | 0.000 | 0.382 | 0.000 | 0.952 | 0.000 |
| 5 | 1.000 | 0.000 | 0.339 | 0.250 | 0.886 | 0.070 |
| 6 | 1.000 | 0.000 | 0.941 | 0.110 | 0.971 | 0.000 |
| 7 | 0.865 | 0.160 | 0.584 | 1.120 | 0.361 | 1.610 |
| 8 | 0.993 | 0.010 | 0.326 | 2.760 | 0.859 | 0.110 |

TABLE IV. RESULTS OF SLOW FLATTENED SENTIMENT DETECTION

| Persona | CUSUM | | EWMA | | SVM | |
|---|---|---|---|---|---|---|
| | F1 | Delay | F1 | Delay | F1 | Delay |
| 1 | 0.964 | 0.000 | 0.417 | 0.670 | 0.299 | 0.400 |
| 2 | 0.983 | 0.030 | 0.636 | 0.180 | 0.667 | 0.210 |
| 3 | 1.000 | 0.000 | 0.222 | 0.000 | 0.629 | 0.270 |
| 4 | 1.000 | 0.000 | 0.644 | 0.120 | 0.838 | 0.060 |
| 5 | 1.000 | 0.000 | 0.771 | 0.060 | 0.771 | 0.140 |
| 6 | 1.000 | 0.000 | 0.987 | 0.000 | 0.987 | 0.000 |
| 7 | 0.875 | 0.050 | 0.707 | 0.320 | 0.338 | 0.220 |
| 8 | 1.000 | 0.000 | 0.324 | 0.200 | 0.667 | 0.290 |

## B. Off-Topic Detection

Off-topic reply detection, identifying responses that gradually drift from the prompt, proved more challenging than detecting sentiment flattening. Performance was assessed across eight personas and three drift speeds (fast, medium, slow), with F1 scores and ROC AUC reported in Tables V–VII.

The GRU model trained on BERT embeddings consistently achieved the highest ROC AUC values (typically >0.95), indicating strong ranking ability. However, F1 scores remained moderate (generally 0.4–0.7), reflecting difficulty in defining clear anomaly thresholds. The strongest performance appeared for Persona 5, a friendly, consistent voice user (F1 = 0.76, AUC = 0.99 in the fast condition), likely due to stable tone and predictable semantic patterns. Lower scores were observed for users with flatter or more variable expression, including Persona 3 (hesitant PM voice responses) and Persona 8 (short, irritable PM replies), where off-topic shifts were subtler and harder to detect.

CUSUM achieved moderate ROC AUC scores for some users (e.g., ~0.8 for Personas 1 and 5) but consistently low F1 scores (<0.3). It struggled to isolate drift points amid natural variability, especially in expressive or voice-based personas. Notably, Persona 8 showed very poor performance under fast drift (F1 = 0.047), due to irregular tone shifts and low expressiveness.

One-Class SVM failed to reliably detect semantic drift across all conditions. F1 scores remained below 0.25 in nearly all cases, and ROC AUC values hovered near random, especially in slow progression scenarios. Persona 6, a highly expressive typed user, scored as low as F1 = 0.013 and AUC = 0.218 under slow drift, suggesting the model could not adapt to gradually evolving topic relevance.

Overall, these results indicate that semantic drift is more complex than affective change to detect. While the GRU model offered promising ranking performance, effective thresholding remains a challenge. Static models such as CUSUM and SVM performed inconsistently without personalized or temporal adjustments, underscoring the need for adaptive and context-sensitive approaches in future systems.

TABLE V. RESULTS OF FAST OFF-TOPIC DETECTION

| Persona | CUSUM | | GRU | | SVM | |
|---|---|---|---|---|---|---|
| | F1 | ROC AUC | F1 | ROC AUC | F1 | ROC AUC |
| 1 | 0.190 | 0.733 | 0.586 | 0.975 | 0.152 | 0.743 |
| 2 | 0.141 | 0.656 | 0.585 | 0.955 | 0.148 | 0.635 |
| 3 | 0.053 | 0.485 | 0.369 | 0.912 | 0.117 | 0.639 |
| 4 | 0.088 | 0.514 | 0.518 | 0.958 | 0.217 | 0.671 |
| 5 | 0.264 | 0.836 | 0.762 | 0.991 | 0.200 | 0.846 |
| 6 | 0.078 | 0.424 | 0.528 | 0.959 | 0.187 | 0.815 |
| 7 | 0.129 | 0.732 | 0.500 | 0.951 | 0.157 | 0.607 |
| 8 | 0.047 | 0.564 | 0.431 | 0.943 | 0.085 | 0.572 |

TABLE VI. RESULTS OF MEDIUM OFF-TOPIC DETECTION

| Persona | CUSUM | | GRU | | SVM | |
|---|---|---|---|---|---|---|
| | F1 | ROC AUC | F1 | ROC AUC | F1 | ROC AUC |
| 1 | 0.261 | 0.830 | 0.660 | 0.975 | 0.152 | 0.744 |
| 2 | 0.176 | 0.699 | 0.549 | 0.953 | 0.168 | 0.663 |
| 3 | 0.106 | 0.544 | 0.406 | 0.894 | 0.168 | 0.598 |
| 4 | 0.097 | 0.709 | 0.568 | 0.965 | 0.138 | 0.639 |
| 5 | 0.413 | 0.941 | 0.725 | 0.985 | 0.230 | 0.863 |
| 6 | 0.096 | 0.631 | 0.517 | 0.933 | 0.178 | 0.616 |
| 7 | 0.197 | 0.753 | 0.545 | 0.953 | 0.184 | 0.602 |
| 8 | 0.102 | 0.513 | 0.507 | 0.932 | 0.164 | 0.518 |

TABLE VII. RESULTS OF SLOW OFF-TOPIC DETECTION

| Persona | CUSUM | | GRU | | SVM | |
|---|---|---|---|---|---|---|
| | F1 | ROC AUC | F1 | ROC AUC | F1 | ROC AUC |
| 1 | 0.286 | 0.864 | 0.637 | 0.972 | 0.152 | 0.720 |
| 2 | 0.202 | 0.872 | 0.478 | 0.952 | 0.081 | 0.618 |
| 3 | 0.120 | 0.623 | 0.331 | 0.875 | 0.071 | 0.436 |
| 4 | 0.168 | 0.801 | 0.469 | 0.956 | 0.020 | 0.250 |
| 5 | 0.507 | 0.945 | 0.642 | 0.975 | 0.199 | 0.770 |
| 6 | 0.095 | 0.642 | 0.405 | 0.941 | 0.013 | 0.218 |
| 7 | 0.170 | 0.768 | 0.517 | 0.956 | 0.174 | 0.582 |
| 8 | 0.072 | 0.508 | 0.384 | 0.903 | 0.051 | 0.344 |

*C. Personalized vs. General Classifiers*

Given the limitations observed in unsupervised and temporal models, particularly in thresholding and sensitivity to gradual semantic drift, experiments were conducted using logistic regression classifiers on BERT embeddings to assess whether richer, contextual representations could improve off-topic detection. While these models are not temporally structured, their semantic granularity offers a stronger signal than handcrafted or statistical features.

Two settings were evaluated: personalized classifiers trained on each individual user's data, and generalized classifiers trained on all but one user via leave-one-user-out evaluation. Results are shown in Tables VIII–X.

Personalized models achieved near-perfect performance across most personas (F1 and ROC AUC > 0.95), with only minor degradation for users with greater variability in tone or structure, such as Persona 6 (high expressiveness) and Persona 3 (time-of-day fluctuation). Generalized classifiers showed clear weaknesses with select personas, with F1 scores dipping below 0.9 and more substantial drops for expressive users. For example, Persona 6 scored only F1 = 0.385 under fast progression and F1 = 0.400 under slow, despite a nearly perfect ROC AUC, illustrating the ongoing challenge of calibration even with rich embeddings.

These results suggest that while contextual models like BERT offer a strong semantic foundation, effective detection still hinges on user-specific calibration. Generalized models may fail to capture the individualized nature of conversational baselines in cognitively sensitive interactions. Incorporating temporal dynamics remains a necessary future step, but these findings confirm the importance of personalization and highlight the value of semantically grounded baselines as a bridge toward more adaptive systems.

TABLE VIII. RESULTS OF PERSONALIZED VS. GENERAL FAST OFF-TOPIC DETECTION

| Persona | General | | Personalized | |
|---|---|---|---|---|
| | F1 | ROC AUC | F1 | ROC AUC |
| 1 | 1.000 | 1.000 | 1.000 | 1.000 |
| 2 | 1.000 | 1.000 | 1.000 | 1.000 |
| 3 | 0.804 | 1.000 | 0.941 | 0.941 |
| 4 | 0.989 | 1.000 | 1.000 | 1.000 |
| 5 | 1.000 | 1.000 | 0.952 | 0.909 |
| 6 | 0.385 | 0.995 | 1.000 | 1.000 |
| 7 | 0.992 | 1.000 | 1.000 | 1.000 |
| 8 | 0.892 | 1.000 | 0.974 | 1.000 |

TABLE IX. RESULTS OF PERSONALIZED VS. GENERAL MEDIUM OFF-TOPIC DETECTION

| Persona | General | | Personalized | |
|---|---|---|---|---|
| | F1 | ROC AUC | F1 | ROC AUC |
| 1 | 1.000 | 1.000 | 1.000 | 1.000 |
| 2 | 1.000 | 1.000 | 1.000 | 1.000 |
| 3 | 0.882 | 1.000 | 1.000 | 1.000 |
| 4 | 0.985 | 1.000 | 0.952 | 0.909 |
| 5 | 1.000 | 1.000 | 1.000 | 1.000 |
| 6 | 0.557 | 0.996 | 1.000 | 1.000 |
| 7 | 0.995 | 1.000 | 1.000 | 1.000 |
| 8 | 0.945 | 1.000 | 1.000 | 1.000 |

TABLE X. RESULTS OF PERSONALIZED VS. GENERAL SLOW OFF-TOPIC DETECTION

| Persona | General | | Personalized | |
|---|---|---|---|---|
| | F1 | ROC AUC | F1 | ROC AUC |
| 1 | 1.000 | 1.000 | 1.000 | 1.000 |
| 2 | 1.000 | 1.000 | 1.000 | 1.000 |
| 3 | 0.868 | 1.000 | 1.000 | 1.000 |
| 4 | 0.986 | 1.000 | 1.000 | 1.000 |
| 5 | 1.000 | 1.000 | 1.000 | 1.000 |
| 6 | 0.400 | 0.999 | 1.000 | 1.000 |
| 7 | 1.000 | 1.000 | 1.000 | 1.000 |
| 8 | 0.937 | 1.000 | 1.000 | 1.000 |

## VI. Discussion

PersonaDrift highlights key challenges and design considerations for conversational monitoring in dementia support contexts. The results demonstrate that detecting behavioral drift in response to digital reminders, such as reduced emotional tone or relevance of replies, requires methods that are both temporally aware and user-specific. Flattened sentiment, characterized by low expressiveness and brevity, was reliably detected using lightweight statistical methods like CUSUM, particularly in personas with stable, typed responses (e.g., Personas 1, 2, 5).

In contrast, off-topic semantic drift was more difficult to detect. These shifts unfold gradually and depend on subtle semantic changes, making them less amenable to statistical or unsupervised models. GRU-based classifiers with BERT embeddings provided strong ranking (ROC AUC) but still struggled with thresholding. One-Class SVMs performed poorly overall, especially under slow progression. These results reinforce the need for richer semantic representations and, ultimately, temporally adaptive models.

Together, the findings suggest that anomaly types must be treated distinctly and underscore the importance of personalization, particularly for semantically complex tasks like off-topic detection.

### A. Personalization as a Core Requirement

Detection performance strongly depended on user-specific modeling. Personalized classifiers trained on individual users' BERT embeddings outperformed generalized ones across all progression speeds, with the performance gap reaching over 0.5 F1 points for expressive or variable personas (e.g., Personas 3, 6, and 8). These results confirm that semantic drift is highly individualized and that generic thresholds cannot reliably detect meaningful change.

Although the benchmark is synthetic, the personas were grounded in real user stories and routine patterns. Each persona reflects different tones, interaction styles, and daily rhythms, such as time-of-day variation (Personas 3 and 8) or emotional variability (Persona 6). These design choices emphasize the importance of modeling user baselines and validate personalization as a core requirement, not an optional enhancement.

### B. Impact of Modality and Expression

Performance also varied with interaction modality and expressiveness. Terse, typed responses (e.g., Persona 1) led to clearer drift signals and better alignment with statistical methods. In contrast, voice-based or emotionally dynamic personas (e.g., Personas 6 and 8) exhibited higher baseline variance, resulting in more false positives and delayed detection, especially for semantic drift.

These findings underscore the importance of modeling not just *what* is said, but *how* it is communicated. Supporting diverse modalities and expressive styles may require adaptive baselines, modality-aware embeddings, or tuning thresholds to a user's natural variance.

### C. Benchmark-First Rationale

Rather than proposing a new model, this work focuses on infrastructure, a benchmark designed to expose challenges in personalized, longitudinal drift detection. While synthetic, PersonaDrift enables systematic evaluation across user types, anomaly forms, and progression speeds, surfacing failure modes such as threshold instability and poor generalization.

Because simulated interactions are more structured than real dialogue, the issues uncovered here (e.g., model sensitivity to variability or progression speed) are likely to be even more severe in real deployments. Addressing these challenges early in development is essential, and the benchmark provides a low-risk environment for testing solutions.

### D. Limitations

*PersonaDrift* offers a structured, simulation-based testbed for evaluating conversational change; it remains a simplified approximation of real-world interactions. The personas are grounded in caregiver interviews and designed to reflect a range of tones, routines, and expressiveness, but they represent relatively stable traits. In practice, PLwD often exhibit dynamic and evolving communication patterns that can shift within and across days. Capturing this kind of temporal variability is beyond the scope of the current implementation.

The benchmark currently includes two anomaly types, affective flattening and off-topic semantic drift, which were selected based on caregiver relevance and modeling feasibility. However, other common features of cognitive decline, such as repetition, lexical retrieval issues, and syntactic simplification, are not yet represented. Including these could provide a more comprehensive picture of linguistic changes over time. We see this as a natural direction for future work, but note that our benchmark's strength lies in controlled, ecologically grounded design rather than raw volume.

We note that data sufficiency varies by detection method, and while our fixed 60-day logs are sufficient for evaluation, systematic sensitivity analysis is left for future work.

Another important limitation is modality. Although some personas are labeled as voice-based, the input data is currently limited to text transcripts. This excludes important multimodal cues such as vocal tone, speech timing, and typing behavior, which can be informative for tracking subtle affective or cognitive shifts [36]. Future versions could incorporate such signals to improve ecological validity.

Additionally, while the benchmark simulates progression over fixed time horizons, it does not currently support adaptive or evolving user profiles. Real-life symptom trajectories may accelerate, plateau, or vary due to interventions, and capturing that kind of fluidity would better reflect the lived experience of users and caregivers.

Lastly, the benchmark does not directly address practical concerns around model interpretability, runtime performance, or deployment constraints, which are important for real-world assistive tools. Nonetheless, by surfacing core challenges like personalization, threshold instability, and anomaly differentiation, *PersonaDrift* provides a strong foundation to guide future research.

*E. Future Directions*

Future work can expand *PersonaDrift* along several dimensions to improve realism and utility. First, additional anomaly types, such as repetition, lexical retrieval issues, or syntactic simplification, would better reflect the linguistic diversity observed in dementia. This expansion would complement the caregiver-driven anomalies prioritized here and further strengthen the benchmark's relevance for both clinical and NLP communities. Second, simulating evolving user traits over longer timelines would allow testing model adaptability to non-stationary behavior.

Multimodal input (e.g., prosody, typing speed, vocal tone) is also critical for cognitive monitoring and could strengthen anomaly detection, especially in expressive or ambiguous cases. On the modeling side, more adaptive approaches are needed. Techniques such as memory-augmented networks, few-shot personalization, and hybrid statistical–neural methods may better capture user-specific drift over time [37], [38], [39].

Finally, extending the generation pipeline to produce task-aligned, privacy-preserving data could support model fine-tuning without relying on sensitive user logs. In parallel, we are conducting usability testing of the reminder system with caregivers, and future work will leverage data collected from these real-world interactions to both validate and refine the benchmark. This will enable direct comparison of ML methods evaluated in PersonaDrift with their performance on authentic conversational data. Together, these additions would bring PersonaDrift closer to deployment-ready evaluation for real-world cognitive support systems.

## VII. CONCLUSION

This work introduces PersonaDrift, a synthetic benchmark for evaluating conversational anomaly detection in cognitively relevant, temporally structured dialogue. Rather than proposing new models, the contribution lies in enabling controlled, reproducible testing of detection methods across user traits and progression patterns.

Flattened sentiment was reliably detected using lightweight statistical methods, while off-topic semantic drift required more complex, personalized approaches. Across all settings, user-specific models consistently outperformed generalized ones, underscoring the importance of personalization in real-world applications.

PersonaDrift serves as a diagnostic tool for identifying limitations in current methods and guiding the development of adaptive, user-sensitive NLP systems. Future extensions may incorporate additional anomaly types, multimodal signals, and in-the-wild deployment to support ethical, personalized cognitive monitoring in home-based care contexts.

ACKNOWLEDGMENT

The authors thank Kelly Beaton and David Black for their invaluable contributions as ongoing lived experience advisors for this study.